\pdfoutput=1
\documentclass[11pt]{article}

\usepackage{EMNLP2023}

\usepackage{times}
\usepackage{latexsym}
\usepackage[T1]{fontenc}
\usepackage[utf8]{inputenc}
\usepackage{microtype}
\usepackage{graphicx}
\usepackage{makecell}
\usepackage{multirow}
\usepackage{inconsolata}
\usepackage{xspace}
\usepackage{diagbox}
\usepackage{booktabs}
\usepackage{wrapfig}
\usepackage{listings}
\usepackage{svg}
\svgpath{{../figures/}} 
\usepackage{rotating}
\usepackage{pdfpages}
\usepackage{enumitem}

\title{Do Language Models Learn about Legal Entity Types during Pretraining?}

\author{Claire Barale \and Michael Rovatsos \\ School of Informatics \\ The University of Edinburgh \\ {\texttt{\{claire.barale,michael.rovatsos\}@ed.ac.uk}} \And Nehal Bhuta \\ School of Law \\ The University of Edinburgh \\ \texttt{nehal.bhuta@ed.ac.uk}}

\begin{document}
\maketitle
\begin{abstract}

Language Models (LMs) have proven their ability to acquire diverse linguistic knowledge during the pretraining phase, potentially serving as a valuable source of incidental supervision for downstream tasks. However, there has been limited research conducted on the retrieval of domain-specific knowledge, and specifically legal knowledge. We propose to explore the task of Entity Typing, serving as a proxy for evaluating legal knowledge as an essential aspect of text comprehension, and a foundational task to numerous downstream legal NLP applications. Through systematic evaluation and analysis and two types of prompting (cloze sentences and QA-based templates) and to clarify the nature of these acquired cues, we compare diverse types and lengths of entities both general and domain-specific entities, semantics or syntax signals, and different LM pretraining corpus (generic and legal-oriented) and architectures (encoder BERT-based and decoder-only with Llama2).
We show that (1) Llama2 performs well on certain entities and exhibits potential for substantial improvement with optimized prompt templates, (2) law-oriented LMs show inconsistent performance, possibly due to variations in their training corpus, (3) LMs demonstrate the ability to type entities even in the case of multi-token entities, (4) all models struggle with entities belonging to sub-domains of the law (5) Llama2 appears to frequently overlook syntactic cues, a shortcoming less present in BERT-based architectures. The code of the experiments is available at \url{https://github.com/clairebarale/probing_legal_entity_types}.

\end{abstract}

\begin{table*}[t]
\centering
\resizebox{\textwidth}{!}{
\begin{tabular}{|l|l|lcccc|} 
\toprule
&\textbf{Pretrained Language Model} & \textbf{Pretraining corpus} & \textbf{\# Parameters} & \textbf{\# Tokens} & \textbf{Corpus size} & \textbf{\# Vocab}\\
\toprule

 \multirow{3}{*}{\rotatebox[origin=c]{90}{\textbf{Legal}}}  &  \textbf{CaseHOLD} \cite{casehold_zhengguha2021} & Harvard Case Law & 110M & 43B & 37 GB  & 32K\\
&    \textbf{Pile of law} \cite{henderson2022pile} & US, Canadian, ECthR &340M& 130B & 256 GB  & 32K\\
&    \textbf{LexLM} \cite{chalkidis-etal-2023-lexfiles} &  US, Canada, EU, UK, India  & 125 & 2T + 256B & 175 GB    & 50K\\       

\midrule
\midrule

 \multirow{7}{*}{\rotatebox[origin=c]{90}{\textbf{Generic}}} & & BookCorpus \cite{zhu2015aligning}  & & -  & 16GB & - \\
&& CC\_news \cite{ccnews_nagel}   && - & 76GB & - \\
&& OpenWebText \cite{radford2019language} && - & 38GB & - \\
&& Stories \cite{trinh2018simple}  && - & 31GB & - \\ 
&\textbf{RoBERTa}\cite{liu2019roberta} &  & 125M & 2T & 160GB & 50K \\
& \textbf{DeBERTa} \cite{he2023debertav} &    & 86M & 2T & 160GB & 128K \\

&\textbf{Llama 2} \cite{touvron2023llama} & Data from publicly available sources & 7B & 2T & - & 32K \\
\bottomrule
\end{tabular}
}
    \caption{Overview of the models used. The table reports the description of the pretraining corpora, the number of parameters, the total number of tokens, the size of the corpus, and the vocabulary size}
    \label{table: plms_table_overview}
\end{table*}

\section{Introduction} \label{intro}

During the initial phase of pretraining, language models (LMs) are exposed to an extensive corpus of textual data, allowing them to acquire the capacity to represent the probabilistic structure of language. In this process, it has been theorized that they incidentally learn various linguistic signals and patterns, both syntactic and semantic. Work by \citet{petroni-etal-2019-language} and subsequent studies \cite{jiang-etal-2020-know} make the hypothesis that a side effect of the pretraining stage is that LMs also learn factual knowledge. On the other hand, \citet{gururangan-etal-2020-dont} research demonstrates the significance of both model pretraining and task-specific pretraining; pretraining a model with a specific focus on a particular task or a limited domain corpus yields notable advantages in enhancing model performance and adaptability. 

Entity typing and extraction are crucial tasks for a range of use cases including named-entity recognition (NER), relation extraction, summarization, structuring raw data, and most specifically in law, legal search, and past cases retrieval. To gain more insights into entity typing and extraction, entity probing tasks have been designed for bidirectional LSTM conditional random field models \cite{AUGENSTEIN}, masked language models \cite{petroni-etal-2019-language, jiang-etal-2020-know} and autoregressive LMs \cite{epure-hennequin-2022-probing}, using GPT-2. 

Conversely, one notable bottleneck of the application of NLP within the legal domain is the lack of resources and annotated datasets. Thereby, it is of particular interest to explore the extent to which LMs, during their pretraining phase, acquire a sufficient \textbf{understanding of legal entities}, serving as a surrogate for legal knowledge. Ultimately, LMs could be exploited as a source of weak and indirect supervision in downstream tasks such as legal NER or question answering (QA), as they constitute a good proxy to use natural text incidentally thanks to their pretraining stage. Indeed, humans do not exclusively rely on exhaustive supervision but instead make use of occasional feedback and learn from incidental signals originating from various sources. This approach holds potential for increased flexibility in terms of entity types, in contrast to supervised methods, and presents an alternative to existing automated annotation extraction approaches \cite{tedeschi-navigli-2022-multinerd,jaromir_annotations} which hold limitations in the set of entity types. It presents several advantages: it does not require human annotation, it can be easily combined with other sources of supervision such as legal knowledge bases, and it would support an open set of entities and user queries. It would offer the advantage of seamless and fast application to new datasets while facilitating transfers of knowledge between datasets and even potentially between different domains. In this paper, we study the \textbf{intersection of entity knowledge and legal knowledge embedded within LMs}, evaluated on a \textbf{\textit{AsyLex}}, a dataset of Canadian Refugee Decisions.

\subsection{Research questions} \label{RQ}
We are interested in evaluating the quality of the entity knowledge learned during pretraining in \textit{ off-the-shelf} LMs, specifically domain-specific entities, such as those pertinent to the legal field. \textbf{How proficient are Language Models at acquiring knowledge about domain-specific entities like legal entities during pretraining?} Can this acquired knowledge be considered sufficiently reliable for tasks such as annotating new datasets or serving as an indirect source of supervision? How does the choice of prompt type impact the results obtained from knowledge queries? To what extent does the variation in acquired knowledge differ across entity types? What categories of factual knowledge can LMs retrieve, and in what instances do they make errors? Does domain-specific pretraining and jurisdiction-specific pretraining enhance the amount of factual knowledge compared to generic pretraining? To what degree does knowledge acquisition in the legal domain overlap with that of general language models?

\subsection{Contributions} \label{contributions}
Differing from the research objectives of \citet{petroni-etal-2019-language}, which focuses on relation extraction, and \citet{chalkidis-etal-2023-lexfiles} which investigates eight distinct legal knowledge probing tasks with a focus on legislation and legal terminology, we focus on \textbf{Legal Entity Types}. To be clear, we ask the LM to predict the entity type, similarly to \citet{epure-hennequin-2022-probing}, and not the actual entity. For example, in the prompt \textit{<Mask> is the capital of Germany}, we expect the answer to be \textit{City} or \textit{Location} and not \textit{Berlin}.

In addition, we adopt a comprehensive interpretation of entity types, aligned with the work of \citet{barale-etal-2023-automated}, encompassing both essential factual knowledge (e.g., location and dates) and more abstract legal concepts, such as the credibility of a claimant and the rationale behind a judgment. Moreover, our approach diverges by allowing longer entities to be masked (Figure \ref{fig:len-token}), where previous work was limited either to single token \cite{petroni-etal-2019-language} or 2-tokens entities \cite{jiang-etal-2020-x, chalkidis-etal-2023-lexfiles}. Where most previous work focuses on masked language modeling objective models (MLM), we introduce the use of autoregressive LM (Llama2) in a zero-shot setting, similar to the approach employed in \citet{epure-hennequin-2022-probing}. 

We make the hypothesis that pretrained LMs inherently contain structured knowledge about specific domains, which could be leveraged to generate incidental training instances. We seek to investigate the depth of a model's knowledge, its nature, and whether it predominantly acquires knowledge from semantic or syntactic cues.

We first conduct in section \ref{pretraining_corpus} an analysis of the pretraining corpus of selected models both generic and legal LMs. We then prompt the LM with two different styles of prompts, cloze text and question-based, for the task of Entity Typing (section \ref{methodo}). After evaluating the experimental results in section \ref{exp_results}, we analyze the type of failure cases (\ref{failure_cases}) to highlight the strengths and weaknesses of the learning process and to draft directions for future work. 

Our contributions are as follows: 
\begin{itemize}[leftmargin=*]
    \item We propose two new experiments on the task of Legal Entity Typing in a zero-shot setting on a large set of entity types: \textbf{\textit{Experiment MLM}} that evaluates generic and legal BERT-based LM on cloze sentences, and \textbf{\textit{Experiment Llama2}} which evaluates Llama2 on QA-style prompts. 
    \item We report the results for both experiments and show that Llama2 exhibits good performance on specific entities and has the potential for improvement with optimized prompts. However, law-oriented LMs display inconsistent results, likely influenced by training corpus variations and struggle with Refugee Law-specific vocabulary.
    \item We propose an in-depth analysis of the failure modes of the models on this task, opening the way for future work.
\end{itemize}


\section{Background and related work} \label{background}

\subsection{Legal NLP and Legal LMs} 
A range of tasks and use cases have been investigated in legal NLP \cite{zhong-etal-2018-legal}, including summarization, information retrieval, and extraction, or question answering. It is worth emphasizing that entity typing is foundational for many of these tasks. 



\begin{figure}
    \includegraphics[width=\columnwidth]{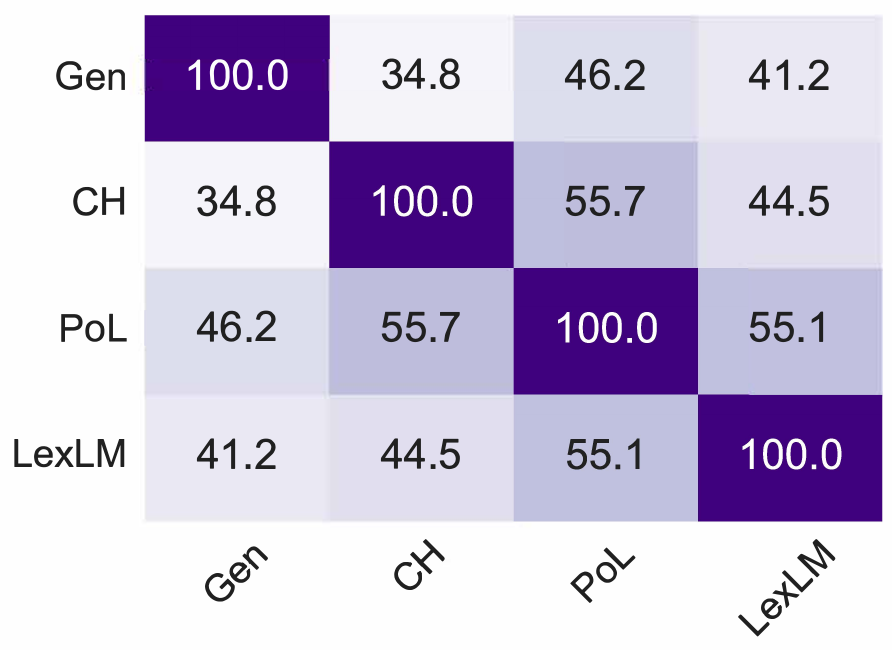} 
\centering
\caption{Vocabulary overlap (\%) between the pretraining corpora. \textit{Gen} stands for \textit{Generic} and is sampled from sources similar to RoBERTa's pretraining corpus, presented in Table \ref{table: plms_table_overview}. Vocabularies are created with the top 10K most frequent tokens in a sample of 50K documents per model}
\label{figure: vocab_overlap}
\end{figure}

The legal domain presents numerous challenges for self-supervised learning, primarily due to the specificity of legal language in contrast to ordinary language. This can lead to ambiguity in contextual meaning (that we aim to assess in this paper), potential implicit meanings, and variations in the significance of a term. A term that may be decisive in a legal context, such as "appeal," might not carry the same weight in a generic domain.

Given these specific challenges and the demonstrated benefits of pretraining LM on legal text to achieve better performance on downstream tasks \cite{barale-etal-2023-automated}, there has been interest in pretraining models on legal texts \cite{zhong-etal-2018-legal, lawformer}. These models typically use an encoder-only architecture based on the BERT architecture: LegalBERT \cite{chalkidis-etal-2020-legalbert}, CaseHOLD \cite{casehold_zhengguha2021}, Pile of Law \cite{henderson2022pile}, and LexLM \cite{chalkidis-etal-2023-lexfiles}, that we use in our first experiment (details in Table \ref{table: plms_table_overview}). To the best of our knowledge, there is no decoder-only legal LM, which is why we limit our second experiment to a Llama2.

\subsection{Probing LMs for Entity Typing}
The idea of latent language representations derived from pre-trained LMs holds promise as a source of structured knowledge. Similar to human learning, LMs accumulate domain-specific and linguistic knowledge, along with the development of general pattern recognition capabilities through their pretraining experiences \cite{brown2020language}. As noted in the introduction, our work follows \citet{petroni-etal-2019-language}'s LAnguage Models Analysis framework (LAMA) and LegalLAMA \cite{chalkidis-etal-2023-lexfiles}. Several probing methods have been investigated \cite{yin-etal-2023-large}, evaluating multilingual extraction \cite{jiang-etal-2020-x} as well as effective prompting for factual knowledge extraction \cite{haviv-etal-2021-bertese, qin-eisner-2021-learning, blevins-etal-2023-prompting}. Various types of tasks have been targeted by these works, including relation extraction, NER, or entity typing \cite{shen-etal-2023-promptner}, our task of interest. Concurrently, there have been efforts to enhance entity typing pipelines, particularly to expand the range of entities beyond traditional categories like location or dates \cite{choi-etal-2018-ultra, dai-etal-2021-ultra} or to entities unseen during training \cite{epure-hennequin-2022-probing, lin-etal-2020-triggerner}, and approaches leveraging supervision from other tasks such as QA \cite{zhang2021entqa}. However, to the best of our knowledge, there has been no work conducted in the legal domain specifically addressing entity typing, and no prior research on entity typing in this domain has made use of prompts in the form of questions.

\section{Pretraining corpus analysis} \label{pretraining_corpus}

\subsection{Vocabulary} \label{vocab_analysis}
To understand the difference between pretraining corpora across LMs, we conduct an exploratory vocabulary analysis inspired by \citet{gururangan-etal-2020-dont} that investigates the impact of domain-specific pretraining on a range of downstream tasks. This preliminary study is destined to clarify and offer insights that will help explain the results of the experiments presented in section \ref{methodo}. We select a total of fifty thousand documents for each language model, perform basic cleaning, tokenize the text, and remove stopwords, which gives us a list of tokens per LM. From this list, we select the most common ten thousand tokens, that constitute the final vocabulary for a given LM. 

For the three legal LMs, as the datasets used for pretraining are directly available, we randomly select the fifty thousand documents. To construct a generic pretraining corpus, we reconstitute a  vocabulary based on the RoBERTa and DeBERTa corpus as indicated in Table \ref{table: plms_table_overview}. As for the other models, we gathered fifty thousand entries, selected proportionally to the size of each corpus. That is to say, we select 5,000 documents from \textit{BookCorpus} which constitutes 10\% of RoBERTa pretraining data, 23,750 entries from \textit{CC\_news}, 11,875 entries from \textit{OpenWebText} (using the open source version: \cite{Gokaslan2019OpenWeb}), and 9,688 entries from \textit{CC\_stories}. Given our limited knowledge of the precise composition of Llama2's pretraining corpus, we propose utilizing the generic vocabularies of RoBERTa and DeBERTa as suitable proxies for our analysis.


\subsection{Vocabulary Overlap} \label{vocab_overlap}
The vocabulary overlap is represented in percentage in the matrix in Figure \ref{figure: vocab_overlap}. As anticipated, the legal LMs exhibit a greater overlap in vocabulary compared to their counterparts with generic training. However, significant disparities emerge among the legal LMs. For example, CaseHOLD shares only 44.5\% of its vocabulary with LexLM. This observation may be attributed to the more extensive and more diverse set of jurisdictions included in the LexLM pretraining corpus. This aligns with the fact that LexLM shows a higher percentage of vocabulary overlap with Pile of Law, which, in contrast to CaseHOLD which is limited to the United States, also includes legal documents from a broader range of jurisdictions.

\begin{figure}
    \centering
    \includegraphics[width=0.9\columnwidth]{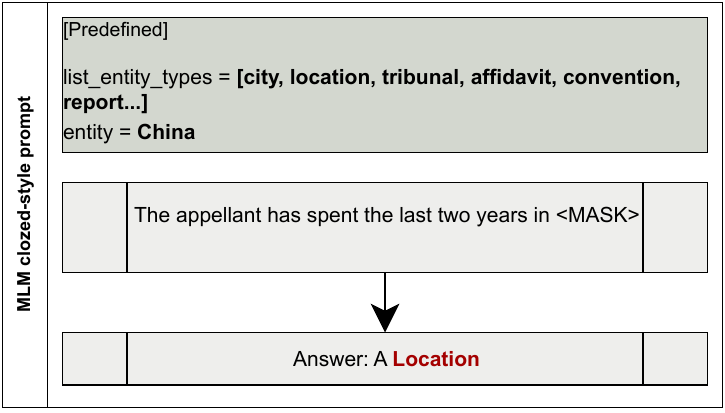}
    \caption{\textit{Experiment MLM} prompt example}
    \label{fig:mlm-prompt}
\end{figure}

\begin{figure}
    \centering
    \includegraphics[width=0.9\columnwidth]{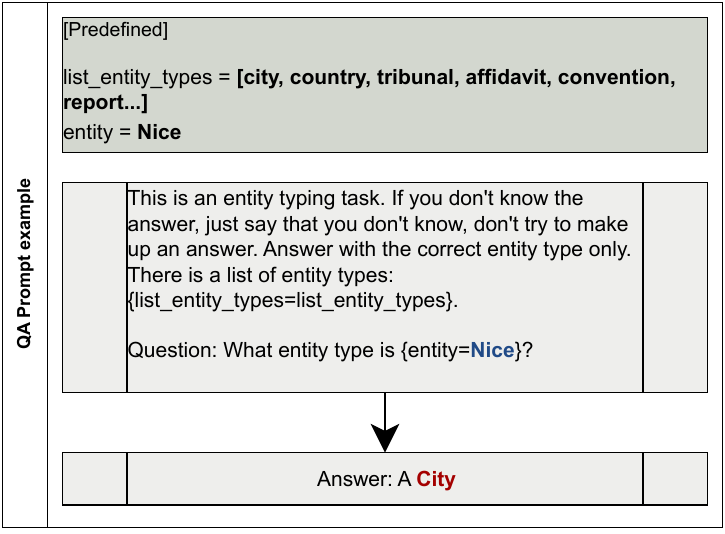}
    \caption{\textit{Experiment Llama2} QA prompt example}
    \label{fig:qa-prompt}
\end{figure}

\begin{figure*}
    \centering
    \includegraphics[width=\textwidth]{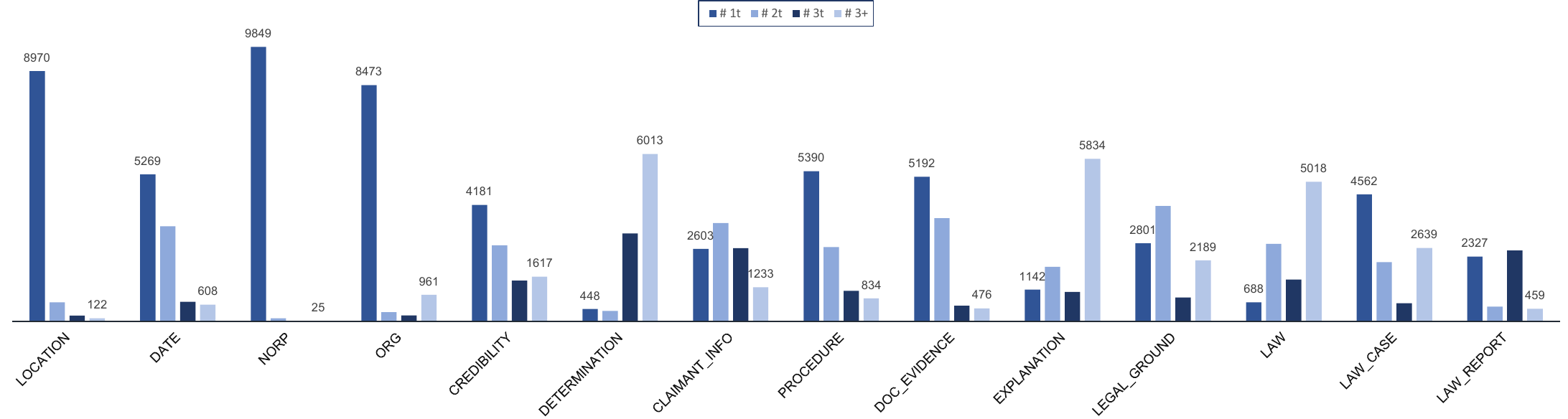}
    \caption{Length of the target masked entities, in number of tokens, for all entity types}
    \label{fig:len-token}
\end{figure*}

\section{Dataset} \label{dataset}
We use \textit{\textbf{AsyLex}}, a dataset of refugee decisions from Canada curated for entity typing and extraction\footnote{\url{https://huggingface.co/datasets/clairebarale/AsyLex}}. This publicly available dataset comprises 19,115 human annotated instances, encompassing 20 distinct categories of entities that hold legal relevance as explained in \citet{barale-etal-2023-automated}. These categories have been identified as categories of interest with the collaboration of legal professionals and experts in the field of refugee law. \textit{\textbf{AsyLex}} comprises 59,112 historical decision documents, spanning from 1996 to 2022. These documents are derived from the online repository of cases of the Canadian Legal Information Institute\footnote{Canlii: \url{https://www.canlii.org/en/}}. 
The documents encompass both initial determinations and subsequent appeals on whether the claimant is granted refugee status or not.
It is important to note that the dataset contains entities of varying generality. Some entities, such as geographical locations, possess broad applicability and could be pertinent to any text (generic entity types: \texttt{location, date, norp}). Others are more specialized within the legal domain, such as procedural steps (generic legal entity types: \texttt{org, law, claimant\_info, procedure, doc\_evidence, law\_case}). Finally, certain entities are highly specific to refugee law, such as the assessment of credibility, which frequently determines the acceptance or rejection of a refugee claim (Refugee Law entity types: \texttt{credibility, determination, explanation, legal\_ground, law\_report}). This diversity in entity scope presents an opportunity for assessing the impact of pretraining, particularly in scenarios where entity types have received various exposures during the pretraining phase.

\subsection{Legal Entity Types} \label{entity_types}
The selection of legal entity types within this dataset is intended to encapsulate characteristics that can reflect similarities among various refugee cases (see Appendix \ref{app_a: entity_description} for an exhaustive description of the types) and for which we have precise gold-standard annotations \cite{barale-etal-2023-automated}. 
The set of 14 entity types is pre-defined and closed for both experiments. To extend the coverage of each entity type and extract specific entities, we use a synonym generator to give synonyms for each of our 14 entity types. As a result, when prompted, the model would have to choose between a total of \textbf{151 entity types}, increasing the difficulty but also the interest of the task. For example, \texttt{location} accepts \textit{city} or \textit{country}. The complete list of synonyms generated per entity type is available in Appendix \ref{app_b: ist_154_ent}. In our evaluation process, we assess predictions across the 14 entity types. For instance, if a prediction is \textit{country}, it will be categorized as \texttt{location} and evaluated against a gold answer that specifies \texttt{location}. 
Contrary to previous work, we do not limit the entities' length to a single token \cite{petroni-etal-2019-language} or to entities spanning only two tokens \cite{jiang-etal-2020-x}. On the contrary, one of the objectives is to use entity types in a broader way for extracting information from text. Thereby we are interested in identifying multi-token entities that have short spans of text and can be longer than two tokens, which is often the case for explaining a decision for instance (entity type: \texttt{explanation}). The length of the entity per entity type is presented in Figure \ref{fig:len-token}, which provides explicit numerical values for both single-token entities and entities longer than three tokens.

\section{Proposed Entity Typing Methodology} \label{methodo}
\subsection{Task description} \label{Task description}
The goal of this task is to classify legal entities mentioned in text documents or sentences into specific types. Legal entities can include various organizations, companies, government bodies (\texttt{org}), or more abstract concepts such as the credibility assessment made in the context of a refugee claim (\texttt{credibility}). The task involves extracting and categorizing these entities based on their attributes or context within the text. As input, we use text documents split by sentences that contain mentions of legal entities. We then categorized legal entity types for each mention found in the input text. 
 
Let \(E\) be the set of possible entity types: \(E = \{e_1, e_2, \ldots, e_n\}\), \(S\) the set of sentences or text segments, \(T\) the set of masked tokens within the sentences and \(P(e_i|t_j, s_k)\) represents the conditional probability that masked token \(t_j\) in sentence \(s_k\) belongs to entity type \(e_i\). The goal is to find the entity type \(e_i\) that maximizes the conditional probability for each masked token \(t_j\) in each sentence \(s_k\):

\[
e_i^* = \arg\max_{e_i \in E} P(e_i|t_j, s_k)
\]

In other words, the objective is to find the entity type that is most likely for each masked token in each sentence.

\subsection{Language models used} \label{LM_used}
For the first experiment, \textit{Experiment MLM} with BERT-based LMs, we experiment with two generic models optimized for MLM, RoBERTa, and DeBERTa, and three legal-oriented LMs (see Table \ref{table: plms_table_overview}). For the second experiment, \textit{Experiment Llama2} we use the open-source model Llama2, optimized for dialogue use cases. Both tasks take the list of entity types as an input argument, making it a multiple-choice task.

\subsection{Cloze prompts with BERT-based models} \label{mlm_prompts}
For the first experimental setting, we use cloze-style prompts that perfectly fit masked language models (MLM). We replace the entities in the sentences with a masked token and use BERT-based models with an MLM objective. Multi-token entities are substituted with a single masked token. If multiple entities appear in the same sentence, only the initial entity occurring in the sentence is considered. The model's answers are limited to the predefined list of 151 entities. We do not provide more context than what is contained in the input sentence. We randomly select ten thousand sentences per entity type, for which we have ground truth annotations (the actual number of prompts after cleaning is given in the column \textit{\# prompts} in Table \ref{tab: mlm_results}). An example of a cloze-style prompt is given in Figure \ref{fig:mlm-prompt}. With this \textit{Experiment MLM}, our objective is to assess whether the models can make predictions about the type of entity to expect based on contextual and syntactic cues. For instance, in the example presented in Figure \ref{fig:mlm-prompt}, we assume that a human reader could deduce from the context that a location is the expected entity to fill the masked portion. Can an LM do the same?

\subsection{QA prompts with Llama2} \label{llama_prompts}
For the second experimental setting, \textit{Experiment Llama2}, we use a template that briefly explains the task to the model and we input the predefined list of 151 entity types. To provide a simple task framing, we prompt the language model according to the following template: "\textit{What entity types is \{entity\}}?", to which the model is asked to answer with the most probable entity type. Because of the format of the prompt, we use a text generation objective with an open-source, state-of-the-art auto-regressive LM, Llama2.  We use the smallest available version of the model (7B parameters, to spare computing resources) and its fine-tuned version Llama2-chat, which is optimized for dialogue use cases. An example of this QA-style prompt is presented in Figure \ref{fig:qa-prompt}. In that experiment, the prompt explicitly mentions the entity, for example, here the question is \textit{"What entity type is Nice?"} which makes it a simpler task compared to the task of \textit{Experiment MLM}. 








\begin{table*}[t]
\centering
\resizebox{\textwidth}{!}{
\begin{tabular}{l||cc|cc|cc|cc|cc||c} 

\toprule
\diagbox{\textbf{Type}}{\textbf{Model}} & \multicolumn{2}{c|}{\textbf{RoBERTa}}& \multicolumn{2}{c|}{\textbf{DeBERTa}}		&\multicolumn{2}{c|}{\textbf{CaseHOLD	}}	&\multicolumn{2}{c|}{\textbf{PoL}}		& \multicolumn{2}{c|}{\textbf{LexLM	}}	& \textbf{\# prompts}\\
\midrule
	&  \textbf{R}&\textbf{F1}&	\textbf{R}&	\textbf{F1	}&\textbf{R}&\textbf{F1	}&\textbf{R}&\textbf{F1}&\textbf{R}& \textbf{F1}	\\
\midrule 

\texttt{LOCATION} &	0.058&	0.110&	0.108&	0.194&	0.070	&0.131&	0.336&	\textbf{0.503}&	0.055&	0.104&	 9,913 \\
\texttt{DATE}&0.036&	0.069	&0.100&	\textbf{0.183}&	0.034&	0.065&	0.025&	0.048&	0.071&	0.133&	 9,442 \\
\texttt{NORP}	&0.037&	0.072&	0.035	&0.067	&0.031&	0.060&	0.032	&0.062&	0.065&	\textbf{0.122}&	 9,986 \\
\texttt{ORG} &	0.088&	\textbf{0.161}&	0.066&	0.123&	0.081&	0.149&	0.018&	0.036&	0.074	&0.138	& 9,947\\  
\texttt{CREDIBILITY} & 0.028	&0.054	&0.026&	0.051&	0.123&	\textbf{0.219}&	0.056&	0.106	&0.028&	0.055&	 9,527 \\
\texttt{DETERMINATION}&	0.384&	\textbf{0.555}	&0.079&	0.147&	0.070&	0.131&	0.142&	0.249&	0.071&	0.133&	 7,242\\ 
\texttt{CLAIMANT\_INFO}	&0.080&	0.149&	0.060&	0.114&	0.081&	\textbf{0.150}&	0.079&	0.147&	0.045&	0.085&	 9,666 \\
\texttt{PROCEDURE}	&0.128&	0.227&	0.080&	0.148&	0.078&	0.145&	0.207&	\textbf{0.344}&	0.228&	0.129	& 9,716 \\
\texttt{DOC\_EVIDENCE}	&0.128	&0.228	&0.125	&0.223	&0.188&	\textbf{0.317}	&0.048&	0.092&	0.056&	0.105&	 9,814\\ 
\texttt{EXPLANATION} &	0.013&	0.026	&0.013	&0.026&	0.009	&0.018	&0.088&	\textbf{0.161}&	0.008&	0.015	& 8,825 \\
\texttt{LEGAL\_GROUND}	&0.029	&0.056&	0.048&	0.091&	0.045&	0.087&	0.041&	0.079&	0.061&	\textbf{0.116}&	 9,640 \\
\texttt{LAW} &0.093	&0.170	&0.203	&0.337	&0.237	&\textbf{0.383}	&0.061	&0.115	&0.066&	0.124	& 9,128\\
\texttt{LAW\_CASE}	&0.079	&0.146&	0.071&	0.133&	0.058&	0.109&	0.106&	\textbf{0.191}&	0.091&	0.167&	 9,290 \\
\texttt{LAW\_REPORT}	&0.057&	0.107	&0.075	&0.140&	0.098&	\textbf{0.178}&	0.087&	0.160&	0.089&	0.164&	 8,601 \\

\bottomrule
\end{tabular}
}
    \caption{Entity type prediction scores in a zero-shot setting, on cloze sentences, measured in Recall and F1 score across 2 generic LMs (RoBERTa and DeBERTa-V3), and 3 legal LMs (CaseHOLD, Pile of Law and LexLM)}
    \label{tab: mlm_results}
\end{table*}

\begin{table}[t]
\centering
\resizebox{0.7\columnwidth}{!}{
\begin{tabular}{l|cc} 

\toprule
\textbf{Type} &  \textbf{R}&\textbf{F1}\\
\midrule 

\texttt{LOCATION} & \textbf{0.956}	&\textbf{0.916 }\\
\texttt{DATE}	&0.730	&0.575\\
\texttt{NORP	}&0.211	&0.118\\
\texttt{ORG}	&0.098	&0.051\\
\texttt{LAW}	&0.100	&0.053\\
\texttt{CREDIBILITY}	&0.219	&0.123\\
\texttt{DETERMINATION	}&0.357	&0.217\\
\texttt{CLAIMANT\_INFO	}	&0.627&0.456\\
\texttt{PROCEDURE	}	&0.259&0.149\\
\texttt{DOC\_EVIDENCE}	&0.653	&0.485\\
\texttt{EXPLANATION} &	0.006	&0.003\\
\texttt{LEGAL\_GROUND}		&0.022&0.011\\
\texttt{LAW\_CASE	}&0.034	&0.017\\
\texttt{LAW\_REPORT	}	&0.048&0.025\\

\bottomrule
\end{tabular}
}
    \caption{Entity type prediction scores in a zero-shot setting, on QA-style prompts, measured in Recall and F1 score, with Llama2, on 10K prompts per entity}
    \label{tab: llama_results}
\end{table}

\section{Experimental Results} \label{exp_results}
We evaluate the results in terms of recall since we want to ensure capturing as many true positives as possible, and F1 score to assess the overall performance on the task. In this section, we compare the results in terms of LM used, length of the input entity, prompt type, and entity type, before conducting an error analysis in the section \ref{failure_cases}. The results of \textit{Experiment MLM} are presented in Table \ref{tab: mlm_results} and the results of \textit{Experiment Llama2} in Table \ref{tab: llama_results}.

\begin{table*}[t]
\centering
\resizebox{\textwidth}{!}{
\begin{tabular}{|c|llllc|} 
\toprule

& \textbf{Error Type} & \textbf{Prompt example} & \textbf{Prediction} & \textbf{Gold} & \textbf{\%}\\
\midrule
\multirow{3}{*}{\rotatebox[origin=c]{90}{\textbf{MLM}}} & Random Prediction &  under <mask> of the Republic of China, they cannot take on a second citizenship & lawsuit & \texttt{law} &70.71\\
&Contextually Accurate &  the applicant has not returned to <mask> since 2008 & employment & \texttt{location} &12.43\\
&Closely Related &my colleague relied on this <mask> in her conclusion & ngo report & \texttt{doc\_evidence} & 16.86\\

\midrule
\midrule

\multirow{4}{*}{\rotatebox[origin=c]{90}{\textbf{Llama2}}}& Random Prediction & What is \textit{Subsection 648}?   & country & \texttt{law} &22.22\\
&Closely Related & What is \textit{vietnamese}?  & country & nationality (\texttt{norp}) &18.52 \\
&False Negative & What is \textit{female claimant}?&female claimant& gender (\texttt{claimant\_info}) &33.33 \\
&Prompt Error & What is \textit{removal order}?  &  It is a type of judicial decision. & \texttt{procedure} &25.93\\

\bottomrule
\end{tabular}}
\caption{Error cases and the ratio of the different error types for both experiments, across all tested models}
\label{table: types of failure cases}
\end{table*}

\begin{table}[t]
\centering
\resizebox{\columnwidth}{!}{
\begin{tabular}{lcccc|c} 

\toprule
& \textbf{Gen} &	\textbf{CH}&	\textbf{PoL}&	\textbf{LexLM}&	\textbf{Llama2}\\
\midrule
\textbf{Generic} &	11.59	&8.51	&\textbf{20.42}&	11.97&	\textbf{63.26}\\
\textbf{Gen Legal}&	17.98&	\textbf{20.89}&	15.40&	12.48&	\textbf{29.52}\\
\textbf{Refugee Law}&	\textbf{10.01}&	5.93&	4.68&	4.92&	\textbf{13.03}\\

\bottomrule
\end{tabular}}
\caption{Entity types prediction scores averaged on 3 groups: generic (\texttt{location, date, norp}), legal entities applicable to most legal domains (Gen Legal: \texttt{org, law, claimant\_info, procedure, doc\_evidence, law\_case}), and legal entities specific to refugee law (Refugee Law: \texttt{credibility, determination, explanation, legal\_ground, law\_report)} \textit{Gen} groups the results of RoBERTa and DeBERTa-v3, \textit{CH} refers to CaseHOLD}
\label{table: results_types-entities}
\end{table}

\subsection{Language Models Comparison} \label{results_lms}
Given the high difficulty of the task, the choice between 151 entity types when accounting for the synonyms list, and the lack of description of the entities and extra context given, it is no surprise that the scores are relatively low. However, the goal of this work is not to reach the best accuracy, but rather to explore where the models succeed or fail. 
On \textit{Experiment MLM}, results are generally lower than in \textit{Experiment Llama2} which is firstly explained by the greater difficulty of the task of \textit{Experiment MLM} and the relative lack of context provided for this task. In this experiment, Pile of Law is the model that performs the best on average, in terms of F1, retrieving 16.36\% of entity types, with 9.47\% in recall. The second best performing model is CaseHOLD with 15.29\% average F1 and 8.58\% average recall. This is despite LexLM's bigger size, LexLM being the model that performs the worst across all, being outperformed by generic LMs RoBERTa and DeBERTa. For all entities except one, the model that achieved the best recall also achieved the best F1, highlighting the consistency in the precision. The only exception is the type \texttt{procedure} for which the best F1 is reached with Pile of Law and the best recall with LexLM. 

\subsection{Single-token vs Multi-token} \label{results_len_tokens}
An interesting point is that we did not impose any restrictions on the length of entities;  the entities that tend to be longer are typically more abstract and closer to a piece of legal common-sense knowledge and reasoning, for example, \texttt{explanation}, \texttt{determination}, \texttt{credibilty} and \texttt{legal ground}. Interestingly the best overall F1 score in \textit{Experiment MLM} is achieved for the type \texttt{determination}, reaching an F1 score of 55.5\% (RoBERTa).
For instance, an entity flagged as \texttt{determination} can as long as: \textit{claimants are not convention refugees and not persons in need of protection}. While the other models achieve lower scores for this entity type, it is to note that the disparity between these relatively lengthy multi-token entities and those that are typically single tokens is not substantial (refer to Figure \ref{fig:len-token}). This may be due to the nature of the task, which may mitigate such disparities compared to tasks like NER where the model has to retrieve the actual entity. In \textit{Experiment Llama2}, shorter entities (that are also the most generic ones) are well recognized (\texttt{location, date}), with also good scores achieved on the types \texttt{determination, claimant\_info, procedure}. Overall for both experiments and certainly due to the nature of the task of entity typing, it seems that the length of the initial entity to categorize does not have an impact on the results. 

\subsection{Prompt Templates Comparison} \label{resuls_prompts}
The scores are on average higher in the \textit{Experiment Llama2} with a total average F1 score of 30.86\% when \textit{Experiment MLM} reaches an average of 14.46\% across all types. However, as noted in the task description (\ref{Task description}) the QA-based experiment is a relatively easier task, making the comparison difficult. 

Based on the predicted entity types, it appears that the template suggested for \textit{Experiment Llama2} is not consistently well comprehended, resulting in a lack of clarity regarding the task. In some cases, it returns not just one entity type, but multiple, leading to incorrect predictions. It seems that, instead of relying on manually crafted prompts and templates, which have been acknowledged to be su-optimal as mentioned by \citet{jiang-etal-2020-x}, there is significant room for improvement in this regard.

\subsection{Entity Types Comparison} \label{resuls_types}
For this evaluation, we categorize the type of entity into three groups: those that can be encountered in any text with the same meaning, those that are commonly found in legal texts, and those that are highly specific to the domain of the dataset, refugee law. The combined results are summarized in Table \ref{table: results_types-entities}. Entities related to refugee law tend to yield the lowest performance across all models and settings. Pile of Law outperforms other models even on generic entities. At the same time, RoBERTa and DeBERTa surpass models specifically trained on legal data for generic legal entities, possibly due to larger exposure and a larger vocabulary.

\subsection{Failure Cases Analysis} \label{failure_cases}
We identify four types of errors across the two experiments: 
\begin{enumerate}[leftmargin=*]
\item Random Prediction: this refers to cases where the predicted entity type is entirely random and unrelated to the context.
\item Contextually Accurate: this describes situations where the predicted entity type is incorrect, but within the context of the sentences, it is plausible in terms of syntax and meaning.
\item Closely Related: instances where the predicted entity class is incorrect, yet it is closely related to the actual gold entity type. For example, it misclassifies a legal ground (which is very precisely one of the 5 reasons for being granted refugee status, see Appendix \ref{app_a: entity_description}) for an explanation of the decision (which is more generic).
\item False Negative, it predicts an entity type that is not in the list of entity types given as input.
\item Prompt Error, if the answer provided deviates from the prompt instruction, we categorize it as incorrect; we consistently consider that an answer with more than five tokens is incorrect, as it signifies that the response extends beyond providing just the entity type.
\end{enumerate}
    
\paragraph {\textit{Experiment MLM} errors} \label{errors_mlm}
To assess the occurrence of error types, we sample 10 errors per entity type per model, i.e. a total of 700 errors for this experiment. Table \ref{table: types of failure cases} presents the findings and an example per error type. There is no instance of a False Negative error; the models never predict an entity type that is not in the input predefined list as we constraint the model to a multiple-choice task, from our pre-defined list of entity types. The most common error is simply an incorrect prediction, with the second most frequent error being the prediction of a closely related entity. This may be due to the choice of categories, some of which express subtle legal nuances. Another positive sign is the presence of more than 10\% of incorrect predictions that are nevertheless accurate in the context of the input sentence. 

\paragraph {\textit{Experiment Llama2} errors} \label{errors_llama2}
Similarly, we sample 10 errors per entity type, i.e. a total of 135 errors. Table \ref{table: types of failure cases} presents the findings. It's worth noting that the use of QA-style prompts leads to a significant number of prompt errors and false negatives, which we believe could be mitigated to some extent by improving the initial prompt template in future work. Additionally, a common misclassification pattern occurs with \texttt{norp} entities, which are always adjectives, but are misclassified as their noun counterparts, as illustrated in the example provided in Table \ref{table: types of failure cases}. Similarly, acronyms for tribunals (e.g., \textit{RPD} for\textit{ Refugee Protection Division}) are classified as \textit{units of length}, an issue that might be rectified by providing more contextual information. Finally, entities like \textit{consistent explanation} are occasionally misclassified as \texttt{explanation} when they should be categorized as a \texttt{credibility} assessment, possibly due to missing adjectives or entity length-related challenges.

\section{Conclusion} \label{conclusion}
Our investigation includes LM selection, input entity length, prompt types, and entity types, in an attempt to understand model strengths and limitations. In summary, our study shows that Llama2 performs best on specific entities and displays potential for improvement with better prompting strategies. However, it also seems that Llama2 repeatedly overlooks syntactic cues. Masked language models mostly appear to be lacking sufficient context within our experimental setup, where they are confronted with a highly challenging task. Law-oriented LMs exhibit varying results, possibly influenced by training corpus differences, and the Pile of Law model shows the best performance on our \textit{AsyLex} dataset. Despite inherent challenges, LMs can accurately identify certain entity types, including multi-token ones, but encounter difficulties with legal sub-domains like Refugee Law.  Future research may explore optimized prompts and few-shot learning strategies. Furthermore, assessing the average precision of the entity type ranking predictions generated by the LM, and conducting experiments on additional datasets, would also be necessary.

\bibliographystyle{acl_natbib}
\bibliography{emnlp2023}

\newpage
\appendix
\section*{Appendix}\label{sec:appendix}

\section{Legal Entity Types Description} \label{app_a: entity_description}

\begin{table*}[h!]
\centering
\resizebox{\textwidth}{!}{
\begin{tabular}{l|p{0.4\linewidth}|p{0.4\linewidth}}

\toprule
\textbf{Type} & \textbf{Description} & \textbf{Examples} \\
\midrule

\texttt{LOCATION}  & cities, countries, regions & "toronto, ontario"\\
\midrule
\texttt{DATE}  & absolute or relative dates or periods & "june, 4th 1996", "two years" \\
\midrule
\texttt{NORP} & adjectives of nationalities, religious, political or ethnic groups or communities & "hutu", "nigerian", "christian" \\
\midrule
\texttt{ORG}  & tribunals, companies, NGOs & "immigration appeal division", "human rights watch" \\
\midrule
\texttt{CREDIBILITY} & mentions of credibility & "lack of evidence", "inconsistencies" \\
\midrule
\texttt{DETERMINATION} & outcome of the decision (accept/reject) & "appeal is dismissed", "not a convention refugee" \\
\midrule
\texttt{CLAIMANT\_INFO} & age, gender, citizenship, occupation & "28 year old", "citizen of Iran", "female" \\
\midrule
\texttt{PROCEDURE} & steps in the claim and legal procedure events & "removal order", "sponsorship for application" \\
\midrule
\texttt{DOC\_EVIDENCE} & pieces of evidence, proofs, supporting documents & "passport", "medical record", "marriage certificate" \\
\midrule
\texttt{EXPLANATION} & reasons given by the panel for the determination & "fear of persecution", "no protection by the state" \\
\midrule
\texttt{LEGAL\_GROUND} & referring to the Convention, refugee status is granted for reasons of race, religion, nationality, membership of a particular social group or political opinion & "homosexual", "christian" \\
\midrule
\texttt{LAW} & citations: legislation and international conventions & "section 1(a) of the convention" \\
\midrule
\texttt{LAW\_CASE} & citations: case law and past decided cases & "xxx v. minister of canada, 1994" \\
\midrule
\texttt{LAW\_REPORT} & country reports written by NGOs or the United Nations & "amnesty international: police and military torture of women in mexico, 2016" \\

\bottomrule

\end{tabular}}
\caption{Pre-defined list of legal entity types}
\label{table: target items}
\end{table*}

\section{Detail of the Extended Entity Types List} \label{app_b: ist_154_ent}

\begin{table*}[h!]
\centering
\resizebox{\textwidth}{!}{
\begin{tabular}{l|p{0.8\linewidth}}

\toprule
\textbf{Type} & \textbf{Extended List}  \\
\midrule

\texttt{LOCATION} & city, country, region, state, province,	area, nation, land, republic, district,	territory, division, zone  \\
\midrule
\texttt{DATE}  & date, day of the month, appointment, particular date, date stamp, time, timestamp, calendar date, schedule\\
\midrule
\texttt{NORP} & nationality, religious community, political group, ethnic groups, community, racial group, party, faction, ideological group, belief community \\
\midrule
\texttt{ORG} & tribunal, firm, ngo,	company, corporation, business,	nonprofit, association,	charity, court,	judicial body\\
\midrule
\texttt{CREDIBILITY} & plausibility, authenticity, integrity, trustworthiness, reliability,	credibility, believability,	credibility, credibleness	\\
\midrule
\texttt{DETERMINATION} & verdict", result, resolution, judgment, approval, denial, decline,	rejection, approval, determination, finding, conclusion, decision, grant, refusal, positive decision, negative decision	\\
\midrule
\texttt{CLAIMANT\_INFO} &  data, employment, resident, national, inhabitant, information, gender, age, citizen,	citizenship, sex, job, occupation, profession \\
\midrule
\texttt{PROCEDURE} &  affidavit, documentary evidence, proof, testimony, exhibit, record, file, paperwork, operation, procedure, legal procedure, legal process, judicial procedure, legal steps,	judicial process \\
\midrule
\texttt{DOC\_EVIDENCE} & proof,	evidence, document,	written document, written evidence,	written proof, written record, written report,	written statement, written testimony, written witness statement \\
\midrule
\texttt{EXPLANATION} &  explanation,	clarification, interpretation \\
\midrule
\texttt{LEGAL\_GROUND} & reason, ground, legal ground, justification, rationale, foundation,	legal basis,	legal justification 	\\
\midrule
\texttt{LAW} &  convention,	international convention,	law,	legislation,	legal code,	treaty,	agreement,	protocol,	statute \\
\midrule
\texttt{LAW\_CASE} & citation,	jurisprudence,	case,	law,	case law,	legal case,	lawsuit,	legal matter,	legal precedent,	judicial decisions,	legal rulings	\\
\midrule
\texttt{LAW\_REPORT} & country	report,	report,	official report,	written report,	ngo report,	national report,	state report,	regional report,	nonprofit report,	non-governmental organization report,	charity report \\

\bottomrule

\end{tabular}}
\caption{Extended pre-defined list of legal entity types (151 types)}
\label{table: extended target items}
\end{table*}

\section{Error Types Detail per LM for \textit{Experiment MLM}}
\label{app_c: error_details}

\begin{table*}[h!]
\centering
\resizebox{\textwidth}{!}{
\begin{tabular}{l|ccccc|c|c}

\toprule
\textbf{Error Type	}&\textbf{RoBERTa	}&\textbf{DeBERTa}	&\textbf{CaseHOLD}	&\textbf{PoL	}&\textbf{LexLM	}&\textbf{\# Total}	& \% \\
\midrule
Random Prediction	&109	&81	&90	&112	&103	&495	&70.71 \\
Contextually Accurate	&7&	27&	25	&16&	12&	87&	12.43\\
Closely Related &24	&32	&25	&12	&25	&118	&16.86\\
False Negative	&-	&-&-	&-&-&-&0.00\\
\midrule
Total	&140&	140&	140&	140&	140&	700&100\\	

\bottomrule

\end{tabular}}
\caption{Error types figures per (number of occurrences) and in percentage for all studied LM}
\label{table:detail error types}
\end{table*}

\end{document}